%% file: main.tex
\documentclass[fleqn,10pt]{wlscirep}
\usepackage{framed,multirow}
\usepackage[utf8]{inputenc}
\usepackage[T1]{fontenc}
\usepackage{url}
\usepackage{hyperref}
\hypersetup{
  colorlinks   = true, 
  urlcolor     = blue, 
  linkcolor    = blue, 
  citecolor   = red 
}
\usepackage{multirow}
\usepackage{multicol}
\usepackage{adjustbox}
\usepackage{lscape}
\usepackage{graphicx}			
\usepackage{subcaption}
\usepackage{longtable}
\usepackage{ltablex,booktabs}
\usepackage{makecell}
\usepackage{amsmath}			
\usepackage{amssymb}			
\usepackage{amsfonts}			
\usepackage{adjustbox}
\usepackage{lscape}
\usepackage{colortbl}
\usepackage{mathtools}

\usepackage{newfloat}
\DeclareFloatingEnvironment[name={Supplementary Figure},fileext=lof]{suppfigure}
\DeclareFloatingEnvironment[name={Supplementary Table},fileext=lof]{suppTable}
\usepackage{lscape}
\usepackage{fltpage}
\usepackage{soul,color}
\usepackage{rotating}
\usepackage{multirow}
 \newcolumntype{C}{>{\centering\arraybackslash}X} 
\begin{document}
\include{titleAuthors}
\include{abstract}
\flushbottom
\maketitle
\flushbottom
\section*{Introduction}

Colorectal cancer (CRC) is one of the most prevalent cancers diagnosed worldwide, contributing to significant mortality rates~\cite{pacal2022efficient}. Recent studies predict that by 2030, more than 1.1 million deaths will be attributed to CRC, with the incidence expected to increase by 60\%~\cite{arnold2017global, karaman2023robust}. Early precursors, such as adenomas and serrated polyps, can help identify individuals at risk of developing CRC. Research indicates that a higher adenoma detection rate (ADR) during colonoscopy is associated with a decreased risk of CRC in subsequent years, hence the detection and removal of benign polyps is crucial in preventing CRC \cite{Bray2018}. Colonoscopy remains the gold standard for polyp detection. This procedure involves inserting a long, flexible tube equipped with a camera through the rectum into the colon, allowing a thorough examination of diseases and abnormalities. However, due to the complex topology of the colon and rectum, navigation can be challenging, requiring skilled endoscopists for effective examination \cite{ali2023multi}. Furthermore, ADR can vary significantly among practitioners; studies have shown adenoma miss rates (AMRs) ranging from 6\% to 41\% \cite{brown2022deep}.

Computer-aided detection (CAD) utilizing machine learning (ML) has been successfully applied in colonoscopy for still images and video data and is currently being used in clinical practice. ML-based models, in particular deep learning (DL), have the potential to improve ADR and reduce the costs associated with polypectomy \cite{ladabaum2023computer}. These computer-aided systems can assist both experienced and beginner endoscopists, ensuring high-quality consistent surveillance for the detection, localization, and segmentation of abnormalities. Regardless of their pathology, significant attention has been given to the automated detection of colon polyps and their characterization as adenomatous or hyperplastic \cite{gan2023application}. However, despite many publications on polyp detection, challenges remain \cite{taha2017automatic}.

In real clinical workflows, polyps are observed across multiple consecutive frames and show significant variation in appearance due to multiple factors, particularly with the presence of blurred images caused by feces, bubbles, or water jets used to clean the colon, which can be misidentified as anomalies \cite{tian2024detecting}. Traditional endoscopy provides continuous video data that enhance temporal detail, significantly influencing polyp detection. The temporal correlations observed in these data are crucial for identifying polyps \cite{xu2024spatio}. Nevertheless, most AI-based polyp recognition models have been trained on static images that do not fully capture the dynamic nature of colonoscopy procedures \cite{krenzer2020endoscopic,dougan2025vim}. 

A significant limitation of existing polyp detection and segmentation methods is the processing of video frames as independent images, ignoring the important temporal relationships between them \cite{jain2025enhancing}. These approaches tend to create a jittering effect due to their high number of false negative predictions~\cite{zheng2019}. This can occur even due minor intensity variations or more lesion like artefacts such as bubbles and other specular reflections in frames. In contrast, leveraging the temporal information of sequential video frames during training can lead to more resilient models, as they incorporate changes in illumination, occlusion, and perspective over time \cite{sharma2023multi}. Moreover, models trained with this spatiotemporal information demonstrate improved generalizability, which is a crucial step toward developing clinically-adoptable and reliable deep learning systems \cite{wang2025improving}. However, research studies that leverage inter-frame relationships for polyp detection and segmentation are still limited in the literature, which could be due to the limited amount of ground truth annotated sequential data.

In order to train temporal-aware methods for polyp recognition tasks, a sequential dataset that captures temporal factors, such as the change in the texture, color, and shape of polyps in successive frames, is crucial for accurate and improved polyp identification \cite{jain2025enhancing}. Furthermore, analyzing the correlation between adjacent frames could mitigate visual aliasing and modeling flaws caused by single-frame methods, while improving temporal coherent predictions \cite{bobrow2023colonoscopy}.

To address these challenges, the Endoscopy Computer Vision Challenge (EndoCV2022) introduces the \textbf{PolypGen 2.0} subchallenge aiming to advance temporal consistency in polyp identification methods and provide a benchmark to evaluate the generalizability of these algorithms through crowd-sourced objectives. PolypGen 2.0 is a multi-center dataset acquired from a diverse population with different clinical settings and endoscopic systems ensuring a diverse data distribution. This challenge focuses on sequence-based evaluation, encouraging the development of methods that exploit temporal dependencies for both polyp detection and segmentation tasks from endoscopy video sequences that are suitable for clinical deployment.

\begin{figure}
    \centering
    \includegraphics[width=\textwidth]{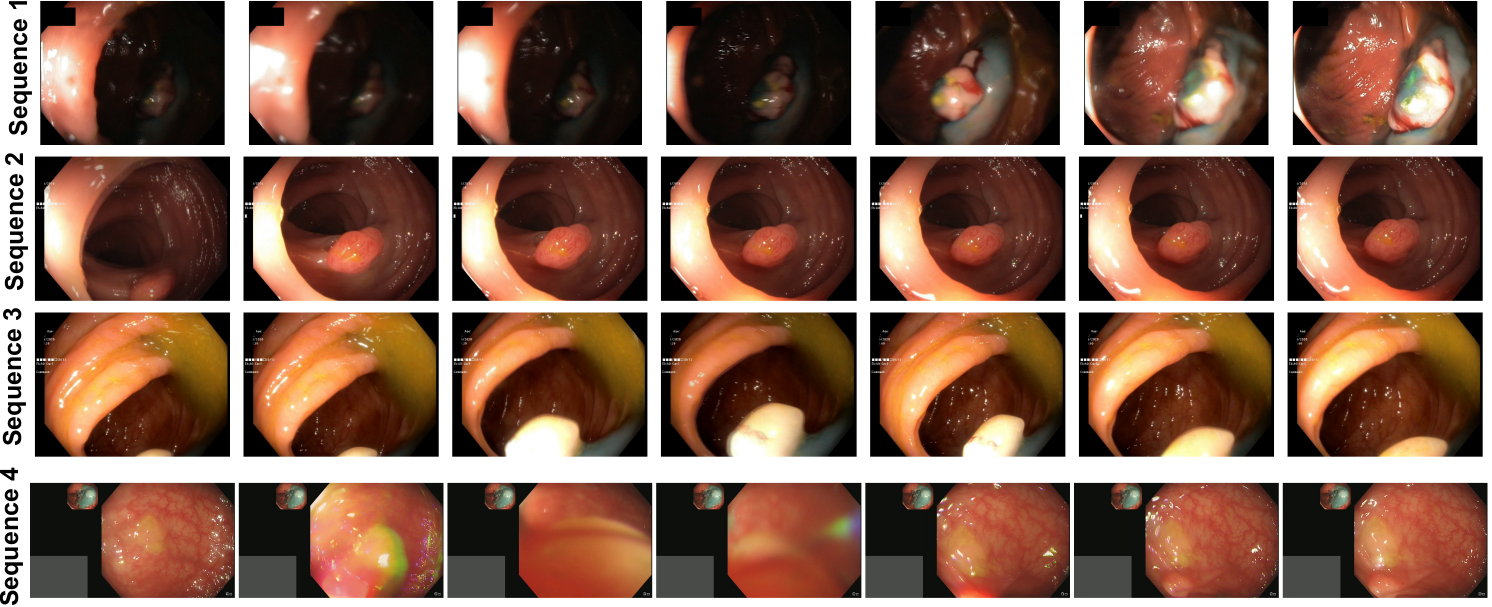}
    \caption{Sequence data samples from a range of patients demonstrate both clear frames and those with artifacts. This highlights the significance of temporal context in accurately interpreting sequence data.}
    \label{fig:Samples}
\end{figure}

\section*{Related work}
In the literature, methods have been developed for real-time analysis, as well as techniques that incorporate temporal information to handle sequence data effectively for both polyp detection and segmentation tasks \cite{zheng2019,ji2022video,GONZALEZBUENOPUYAL2022102625}. This study concentrates on methods from the literature that make use of temporal information to enhance sequence data analysis. Generally, \textit{Polyp detection} methods can perform different tasks such as polyp localization, tracking, and classification\cite{li2019clu,ali2024assessing}. While \textit{Polyp Segmentation} is the process of separating the target polyp from its background in the image. This section thoroughly reviews recent deep learning methods for polyp detection and segmentation that are available in the literature. In each subsection below, we discuss traditional CNN methods, focusing on real-time analysis and methods incorporating temporal information to deal with sequence data. 

\subsection*{Polyp detection and localization:}
The main goal of a detection method is to ensure whether the detected region belongs to a predefined class. This can provide a second opinion and guidance to clinicians, in addition to counting polyps to assess the patient’s condition. In the literature, various papers propose different polyp detection methods, including single-stage, multi-stage, and ensemble models. These methods, which are each designed to resolve specific challenges in detection tasks, provide distinctive advantages. Single-stage models provide faster inference times, while multi-stage models typically enhance accuracy through more refined processing. Ensemble models, on the other hand, leverage the strengths of multiple methods to improve overall performance. To study the temporal information from sequence data, some methods include Recurrent Neural Networks (RNN)\cite{medsker2001recurrent} to their detection model or explore relationships between the frames.

\subsubsection*{Traditional CNN methods}
 
Both the single-stage and multi-stage detectors have been employed by different methods in the literature for polyp detection.  Amer et al.\cite{ amer2024generalized} proposed a model named EDF-YOLOv8 that enhances the performance of the YOLOv8 by employing deformable convolution in the bottleneck and incorporating the Exponential Linear Unit in the model. The model was trained and tested on two different datasets to ensure its overall generalizability. Fu et al.\cite{fu2024d2polyp} proposed a real-time polyp detection and diagnosis network named D2 polyp-Net that utilizes a cross-modal pairwise training strategy and a space-guided localization framework. The space-guided architecture provides a short-range information flow that connects shallow-spatial and deep-semantic features, allowing D2 polyp-Net to accomplish more precise localization and provide effective spatial information for polyp classification. The classification performance is enhanced by the cross-modal pairwise training strategy, which effectively reduces the classification difference for different modalities. Also, Shin et al. \cite{shin2018automatic} integrates the deep learning model Inception-Resnet with the Faster R-CNN accompanied by different image augmentation strategies to detect polyps. Moreover, the method incorporates two post-learning methods for false positive learning and offline learning. However, those methods do not take into account the temporal information encoded in the endoscopic video sequences.

\subsubsection*{Methods incorporating temporal information}
Some work in the literature incorporates temporal dependency in their computer-aided detection methods for colonoscopy data, in particular, polyps. Goa et al.\cite{gao2023tpnet} suggested a model called TPNet, which is a weakly supervised way to detect polyps in videos. It includes a bi-directional LSTM-based temporal encoder and a self-attention mechanism that captures the patterns and temporal variations within polyp video segments. Jiang et al.\cite{jiang2023yona} suggest an end-to-end framework that utilizes the details of one adjacent frame with reference to the current frame for polyp detection with a model named YONA (You Only Need One Adjacent Reference Frame). The model aligns the channel patterns of two frames based on their foreground similarity while dynamically aligning the background based on inter-frame differences to eliminate invalid features caused by drastic spatial jitters. The model also employs a cross-frame box-assisted contrastive learning module to enhance its ability to differentiate between polyps and backgrounds by utilizing box annotations. Additionally, Wu et al.\cite{ wu2021multi} introduce Spatial-Temporal Feature Transformation (STFT), which effectively aggregates features from neighborhood frames and provides more accurate predictions by mitigating inter-frame variations in the camera-moving situation with feature alignment in proposal-guided deformable convolutions. The model is equipped with a new channel-aware attention module that effectively integrates features while maintaining a good  balance between efficiency and expressiveness.

\subsection*{Polyp segmentation:}
The segmentation procedure in medical imaging involves the separation of the target organ or disease from its background in the image. Polyp segmentation deep learning methods fall into the following categories: Models focusing on local representation using CNNs, encoder-decoder architecture and Transformers, which are adept at capturing global context \cite{wan2025umf}
Despite the majority of methods relying on widely-used UNet-based encoder-decoder architectures, they employ single frames for both training and testing. However, some methods also consider temporal information to improve segmentation output \cite{klymenko2021butterfly}.
 
\subsubsection*{Methods involving traditional CNN methods}

In the literature, segmentation models emphasize ensemble methods that integrate various strategies for feature extraction. Tomar et al. \cite{tomar2022tganet} suggest a text-guided polyp segmentation method called \textit{TGANet}. It uses size- and polyp number-related features to build a text-guided attention model that better represents characteristics. The decoder blocks learn an auxiliary task alongside the primary task to supplement the size-based and number-based feature representations. Moreover, Jain et al. \cite{ jain2023coinnet} proposed \textit{CoInNet} that includes a feature extraction system that utilizes the advantages of convolution and involution strategies to emphasize key regions in endoscopic images by analyzing the interrelations among several feature maps using a statistical feature attention unit. Additionally, the authors present an anomaly boundary approximation module that utilizes recursively fed feature fusion for refined segmentation results. An encoder-decoder architecture with a residual axial attention module for feature fusion named \textit{ColonFormer} is presented by Duc et al. \cite{ duc2022colonformer}. The encoder is a lightweight design that utilizes transformers to model global semantic relationships on several scales. On the other hand, the decoder is a hierarchical network architecture intended to capture multi-level features and enhance feature representation. Zhou et al. \cite{zhou2023cross} handled the variety of sizes and shapes of polyps through a cross-level feature aggregation network (\textit{CFA-Net}). The model develops a two-stream architecture-based segmentation network to leverage hierarchical semantic information from cross-level characteristics. Additionally, a Cross-level Feature Fusion (CFF) module is designed to merge neighboring features from several levels, effectively capturing cross-level and multi-scale information to address scale variations of polyps. For finer segmentation maps that improve the hierarchical features, an aggregated module  is introduced to integrate border information into the segmentation network. Furthermore, Toman et al. \cite{toman2025espnet} present an ESPNet for polyp segmentation, which is a transformer-based model that enhances the interaction between local features using a proposed feature decoding and fusion technique with polyp edge features and employing the global contextual information. 

\subsubsection*{Methods incorporating temporal information}
Several studies for Video Polyp Segmentation in the literature incorporate temporal information in segmentation tasks. Puyal et al.  \cite{puyal2022polyp,puyal2020endoscopic} proposed a hybrid 2D/3D convolutional neural network architecture for polyp segmentation. The model takes advantage of the 2D architecture to learn spatial information and provides the chance of applying transfer learning for the still images. Additionally, the 3D-CNN integrates the temporal correlation in the final segmentation results by learning spatio-temporal information. Moreover, Li et al. \cite{li2022tccnet} presents a novel model named  Temporally Consistent Context-Free Network (TCCNet) for polyp video segmentation by designing a Sequence-Corrected Reverse Attention (SC-RA) module and a Propagation-Corrected Reverse Attention module (PC-RA)  that save the predictions temporally constant for successive frames.  Moreover, Zhao et al. \cite{zhao2022semi} suggests a spatial-temporal attention network. The model consists of the Temporal Local Context Attention (TLCA) module and Proximity Frame Time-Space Attention (PFTSA) module. The TLCA module is responsible for improving the prediction from the current frame using the prediction results from the surrounding frames. Additionally, the PFTSA module uses a hybrid transformer architecture to capture long-range dependencies in both time and space.Peng Ji et al. \cite{ji2022video} developed a baseline for polyp segmentation from videos, termed \textit{PNS+}, which comprises three components: a global encoder, a local encoder, and normalized self-attention blocks. To derive long- and short-term spatiotemporal representations, the global and local encoders process an anchor frame together with several subsequent frames, which are subsequently refined by two normalized self-attention blocks. The model was evaluated using the SUN-SEG dataset, comprising 158,690 colonoscopy video frames, and shown exceptional performance. Gao et al. \cite{ gao2023tpnet } propose TPNet, which incorporates a temporal encoder established on bi-directional LSTM and a self-attention mechanism that successfully captures the temporal dynamics and detailed patterns within polyp video segments to enhance accuracy. Moreover, Lu et al. \cite{lu2024diff} introduce an innovative diffusion-based network for the video polyp segmentation challenge, referred to as Diff-VPS. The approach integrates multitask supervision with diffusion models to facilitate pixel-by-pixel segmentation. This combines the high-level contextual details achieved by mutual classification and detection tasks. A Temporal Reasoning Module (TRM) is developed to investigate temporal dependency by reasoning and reconstructing the target frame from preceding frames. Furthermore, Wang et al. \cite{ wang2024tsdetector} presented a temporal-spatial self-correction detector (TSdetector) that combines reliability learning at the spatial level with consistency learning at the temporal level to constantly detect polyps from sequence frames. A global temporal-aware convolution is used by the TSdetector to help the convolutional kernel concentrate on features that are the same across sequence frames. It also uses a hierarchical queue integration technique to combine features from different times using a progressive accumulation method. Additionally, Biffi et al. \cite{biffi2025temporal} present a TCN-based approach called ColonTCN, which employs stacked customized temporal convolutional blocks that utilize double-dilated residual convolutions and regularization. The model is developed to effectively capture prolonged temporal dependencies for the temporal segmentation of colonoscopy videos. Zhao et al. \cite{zhao2025efficient} presented a Deformable Alignment and Local Attention (DALA) method to mitigate the background motion in endoscopes that hinders polyp variation segmentation. The approach utilizes a shared encoder that concurrently encodes the features represented by the corresponding video frames. A deformable convolution module, termed Multi-Scale Frame Alignment (MSFA), is presented to estimate motion between anchors and reference frames. The MSFA is designed to accommodate the diverse scale of polyps observed from various perspectives and motions during the colonoscopy procedure. Furthermore, they utilize Local Attention (LA) to consolidate the aligned features, resulting in more accurate spatial-temporal feature representations.

\nopagebreak
\section*{Material and methods}

\subsection*{EndoCV2022 Challenge}

\subsubsection*{Challenge setup}

A challenge website with an automated docker system for metric-based ranking procedures was set up (see \url{https://endocv2022.grand-challenge.org/}). Participants in the challenge were required to execute inference on our cloud-based system, which utilized the NVIDIA Tesla V100 GPU, and were provided with a test dataset along with instructions for direct GPU usage without downloading the data during the initial two rounds. Round 3 was introduced to evaluate the models trained by participants on an additional unseen sequence dataset by the organizers and to analyze the methods for fairness reporting and experiment-based hypotheses as learning opportunities. The challenge comprised three rounds, utilizing test frames derived from previously undisclosed patient data to avoid data leakage. 

\subsubsection*{Ethical approval and privacy aspects of the data}

The EndoCV2022 data was collected from six distinct centers across five countries: the Egypt, France, Italy, Norway, and UK. Each center managed the ethical, legal, and privacy aspects of its own data. All data were acquired with informed consent and were exempted by the local authority. All institutions required the necessary approvals. The imaging data collected received approval from IDRCB for Ambroise Paré Hospital (Paris, France) (IDRCB: 2019-A01602-55); the institutional research ethics committee at John Radcliffe Hospital (Oxford, UK) sanctioned the collection and utilization of the data under REC Ref: 16/YH/0247, while additional images obtained from other centers were authorized by the institutional data inspectorate. It is important to emphasize that no tissue samples were utilized. All imaging data utilized in this investigation were gathered and completely anonymized according to the General Data Protection Regulation (GDPR) and the Declaration of Helsinki. All procedures were conducted in accordance with the applicable rules and regulations. 

\subsubsection*{Evaluation and ranking}
For the \textbf{Detection task}, the standard metrics along with the generalization metrics are employed as follows: \textit{\textbf{Intersection-Over-Union}} (IoU) quantifies the overlap between two bounding boxes, A and B, as the ratio of the target mask to the anticipated output. $\text{IoU(A,B)} =\frac{A \cap B} {A \cup B}$.
Here, $\cap$ represents the intersection and $\cup$ represents the union. \textit{\textbf{Average Precision}} (AP) is computed as the Area Under Curve (AUC) of the precision-recall curve of the detection sampled at all unique recall values ($r1, r2, ...$) whenever the maximum precision value drops. The mathematical formulation is given by: $\mathrm{AP} = \sum_n{\left\{\left(r_{n+1}-r_{n}\right)p_{\mathrm{interp}}(r_{n+1})\right\}}$. 
Here, $p_{\mathrm{interp}}(r_{n+1}) =\underset{\tilde{r}\ge r_{n+1}}{\max}p(\tilde{r})$. Here, $p(r_n)$ indicates the precision metric at a specific recall level. This definition guarantees a monotonically decreasing precision. AP was computed as an average of APs from $0.50$  to $0.95$ with an increment of 0.05. Additionally, we have calculated
AP\textsubscript{\textit{small}}, AP\textsubscript{\textit{medium}}, AP\textsubscript{\textit{large}}.  

For the \textbf{Segmentation task}, widely accepted computer vision metrics were used: \textit{\textbf{Dice Coefficient Score}} ($DSC = \frac{2 \cdot tp} {2 \cdot tp + fp + fn}$), \textit{\textbf{Jaccard Coefficient}} ($JC= \frac{tp} {tp + fp + fn}$), \textit{\textbf{precision}} ($p=\frac{tp} {tp + fp}$), \textit{\textbf{recall}} ($r= \frac{tp} {tp + fn}$), \textit{\textbf{overall accuracy}} \textit{($Acc = {\frac{tp + tn} {tp + tn + fp + fn}}$ )}, and \textit{\textbf{F2}} ($=\frac{5p \times r} {4p + r}$). Here,  \textit{tp}, \textit{fp}, \textit{tn}, and fn represent true positives, false positives, true negatives, and false negatives, respectively.  A commonly employed segmentation measurement that relies on the distance between two point sets, ground truth (G) and estimated or predicted (E) pixels, was used. The metric is known as an average \textit{\textbf{Hausdorff distance}} ($H_{d}$) and is formulated as $ H_{d}(G, E) = \bigg(\frac{1}{G} \sum_{g\in G} \min_{e\in E} d(g, e) + \frac{1}{E} \sum_{e\in E} \min_{g\in G} d (g, e)\bigg)/2$. $H_{d}$ is normalized between 0 and 1 by dividing it by the maximum value for a given test set. Thus, $1 - H_{d}$ can be regarded as the interval wherein increased values indicate a reduced distance between the ground truth and the estimated segmentation outline.

For team ranking, we compute the overall ranking (R$_{overall}$) for each task taking into account the algorithmic performance on one main metric (R$_{algo.}$) and the inference time ranking (R$_{time}$). For example, for the detection task R$_{overall}$ is defined as the weighted sum between the mean Average Precision (mAP) for intersection threshold 0.50 to 0.95 with an interval of 0.05 and the inference time.  Similarly, for the segmentation task weighted sum of Dice Similarity Coefficient (DSC) and the inference time is used. We use higher weight (3/4th) for the algorithmic ranking.

\subsection*{Dataset}

The EndoCV2022 challenge addresses the generalisability of polyp detection and segmentation tasks from a sequence of endoscopic frames. The colonoscopy video frames utilised in the challenge are collected from six different centres (C1: Ambroise Paré Hospital, France; C2/C3: Centro Riferimento Oncologico and IOV, Italy; C4: Oslo University Hospital, Norway; and C5: John Radcliffe Hospital, UK; and C6: University of Alexandra, Egypt), including HD and Ultra HD. The dataset comprises 46 sequences, totaling 3,290 annotated frames. Each sequence ranges from 12 to 250 frames and includes only the polyp class with total annotated polyp labels of 3014. It consists of 659 (20\%) empty frames (no polyp instance) and 2631 (80\%) frames with polyp instances. For majority of frames, the number of instances of polyp in a single frame is one with only 319 frames with more than one polyp instances. The highest number of polyp instances is five but in just one frame followed by 7 frame with four and 47 with three, and 264 with two instances. The variability in polyp occurrence mimic real-world colonoscopy data. Figure \ref{fig:Samples} shows samples of different sequences consisting of five frames from various patients. These sequences illustrate both clear samples (i.e. Sequence 1) and instances of blurring due to obstacles or occlusions (i.e. Sequence 2, Sequence 3 and Sequence 4). This variation highlights the significance of temporal context by examining the relationship among frames for improved detection. The polyp labels, with precise delineation of polyp boundaries (pixel level for segmentation tasks and bounding boxes for detection tasks), were verified by six senior gastroenterologists and consist of both small and large polyps. 

\begin{figure}
    \centering
    \includegraphics[scale=0.9]{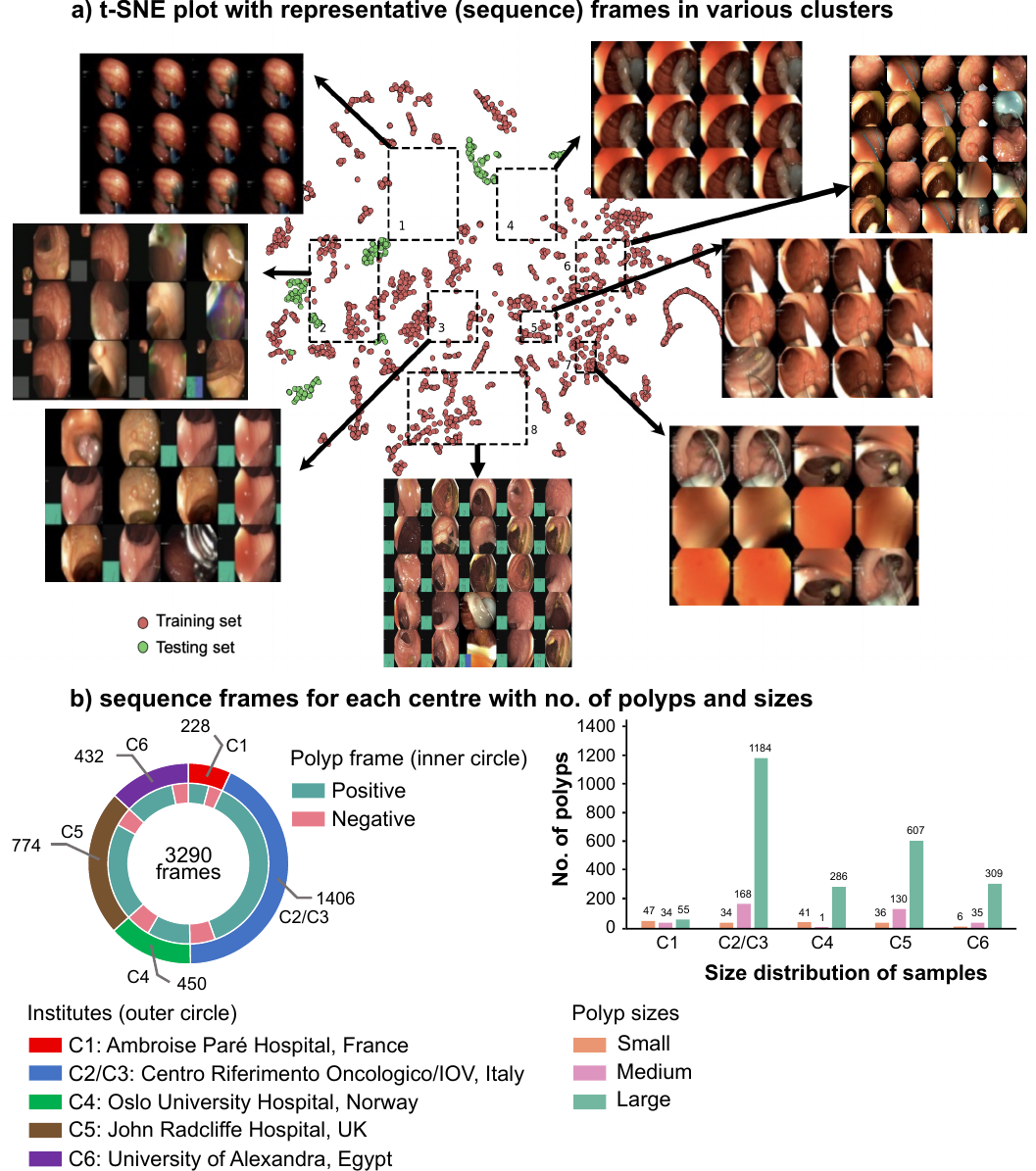}
    \caption{ (top) A $t$-SNE plot for the distribution of training and testing set from the six different centers. Each point represents a frame from the multi-center dataset, color-coded according to the center originated from. The distribution highlight the challenges of obtaining a cross-center generalization and the significance of model robustness on various data from different resources. (b) Illustrates the dataset variability in terms of the number of frames from each centre and the number of polyps in these sequences.}
    \label{fig:tsne}
\end{figure}

The test data was curated to include distinct additional nine video sequences that consist of over 360 frames from multiple centers, providing a robust generalizability test of algorithms. The total number of frames with polyp is 248 (~69\%) and non-polyp (empty) frame is 112 (~31\%). The total number of polyp instances in test set is 255. Modality, population, endoscope model or manufacturer, and polyp size comprise the test categories.  A $t$-SNE plot for the distribution of training and testing set is presented in Figure \ref{fig:tsne} (a). Also, the positive and negative frame distributions for each center is provided in Figure \ref{fig:tsne} (b) with each center-specific polyp size distribution in the histogram (on the right). It could be observed that C2/C3 has higher number of large polyps compared to any other center. The reason behind combining C2 and C3 was their appearance of the polyp samples being very close to each other. 

\subsection*{Annotation Protocol:}
A team of three experienced researchers employed the online annotation tool Labelbox (https://labelbox.com)  for the annotation procedure. Each annotation was cross-validated by the team and the center expert to ensure the precise segmentation of polyp boundaries. A binary review process was conducted independently, with at least one senior gastroenterologist assigned. To facilitate the manual annotation of polyps, the following protocols were developed: 
\begin{itemize}
    \item Clear raised polyps: Boundary pixels must include just the protruding areas. We implemented precautions during the delineation of the standard colonic folds.
    \item Inked polyp regions: Only a segment of the delineation of the non-inked accessible object.
    \item Polyps with instrument parts: Annotations must exclude the instrument and should be meticulously defined.
    \item Pedunculated polyps:  Annotation must include all raised regions unless they are located on the colon fold.
    \item Flat polyps: Expanding the regions found with flat polyps before manual delineation and consulting a qualified specialist if required.
    
\end{itemize}

\subsection*{Challenge Tasks}
EndoCV2022 included two tasks (1) detection and localisation task and (2) pixel-level segmentation task,  generalizability assessments were implemented for both tasks. Sequence frames with manually annotated ground truth polyp labels and their corresponding bounding box locations (origin, height, and width) were provided to participants for the detection task. In order to predict the "polyp" class label, bounding box coordinates (origin, height, and width), and confidence scores for localization, participants were required to train their model. For the semantic segmentation task, expert-provided pixel-level segmentation ground truth that included the same data as provided for the detection task. The participants were assigned with acquiring a binary map prediction for each pixel that was as close to the ground truth as possible, with a value of zero for the background and one for the polyp. The generalizability of the methods that were devised was rigorously evaluated for both of these challenge tasks.

\subsection*{Method summary for participating teams}
\begin{figure*}[t!]
\centering
\includegraphics[width=\textwidth]{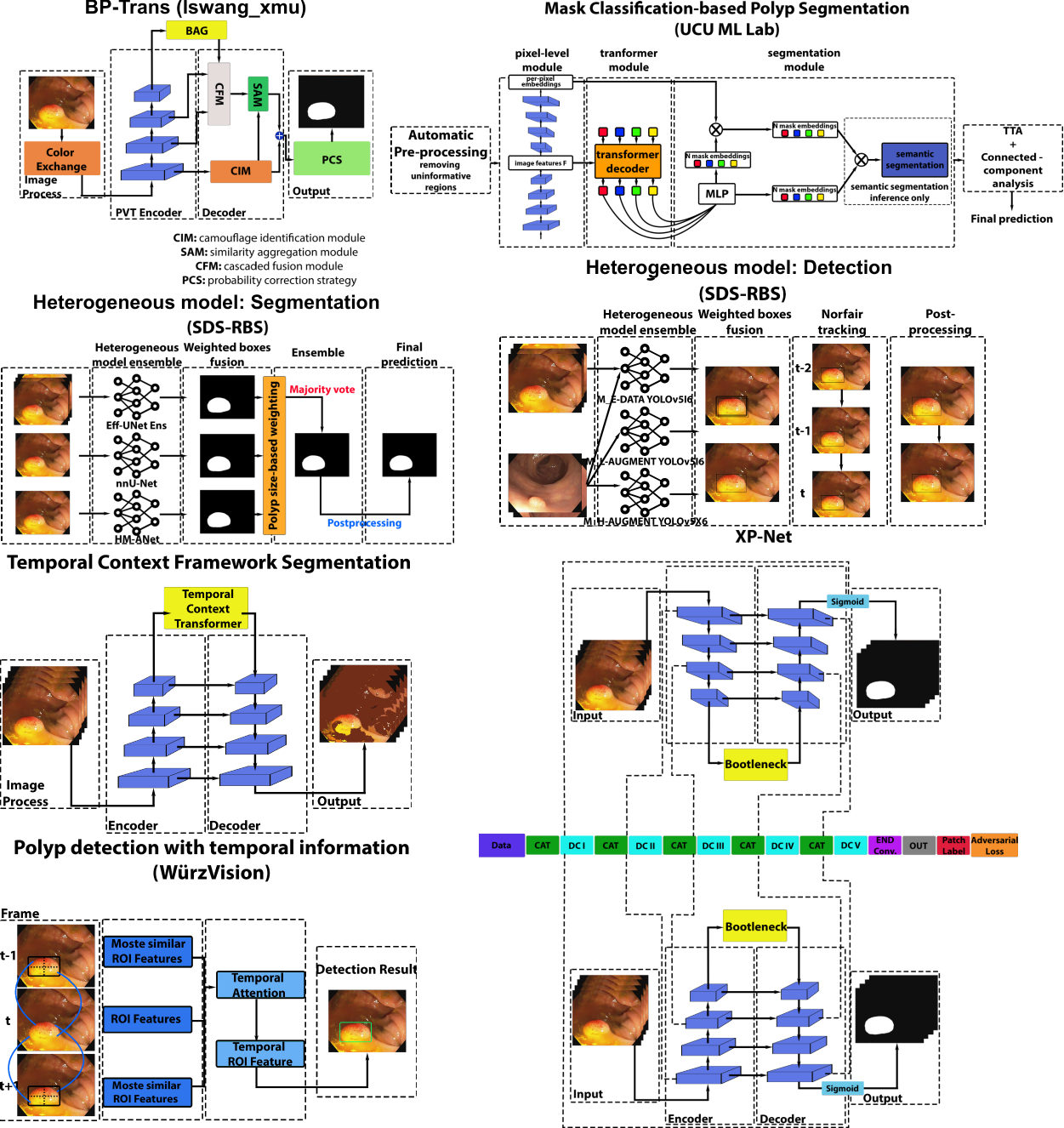}
	\caption{Different architectures proposed by various EndoCV2022 participating teams. Each method takes an input image, and the output prediction is generated directly or through an ensemble of networks. Table 1 presents the description of the backbone and features of various networks. Each output prediction for the detection process consists of a bounding box prediction with the class label "polyp", whereas for segmentation, it involves pixel-wise classification, with polyp pixels assigned label 1 and background pixels assigned label 0.}
    \label{fig:method_summary}
\end{figure*}

In this section, a detailed description of the model proposed by the top participating teams for the EndoCV2022 challenge. Each of these teams has participated in either detection or segmentation tasks or both. A summary figure for the models presented by the team is provided in Fig. \ref{fig:method_summary}. 

%
\subsection*{Team: Arrah\_htic}
For solving the segmentation task, team Arrah\_htic  employed a master-student architecture with two separate teacher models~\cite{DBLP:conf/isbi/BRGCSS22}. The individual models were based on U-Net architectures in combination with a multiscale attention mechanism. An Effective Pyramidal Squeeze Attention (EPSA) block as the first encoder was used in the U-Net architecture for this purpose.  This allows capturing spatial features at multiple scales and to focus on crucial areas. For the student model, separable filters were included in the architecture to reduce the number of learnable parameters. Therefore, the student model has lower computational costs to enable real-time performance. Hierarchical knowledge distillation techniques were used to transfer the complementary knowledge of both teachers to the student network. During training, the challenge data was split into train and test data and augmented using rotation, flipping, and perspective transforms as well as color augmentations to simulate brightness, contrast, and hue variations. The team used the segmentation mask outputs to report their detection results for this challenge. We have not included this team for detection task ranking. 

\subsection*{Team: He\_HIK}

The He\_HIK team improved the Space-Time Correspondence Networks (STCN) model with a semi-supervised learning method to enhance the model’s generalization and the polyp segmentation task ~\cite{DBLP:conf/isbi/HeHSZWW22}. The model utilizes ResNet50 and ResNet18 as the backbone network to build a key encoder and value encoder, respectively. The key encoder encodes the images into the key feature space, while the value encoder encodes both the images and masks into the feature space. To capture temporal information in the video sequences, the frame is encoded in the key feature space, and then the similarity to the key features of the previous frame stored in the memory bank is calculated. The training procedure starts with pre-training a single frame segmentation network (SFSN). The SFSN parameters will be the STCN model's pre-training parameters. The final inference starts with the SFSN model predicting the first frame and the STCN tracking the masks in the following frames.

\subsection*{Team: iMED}
The iMED team used two different architectures and methods to solve both the endoscopy artefact detection and segmentation tasks~\cite{DBLP:conf/isbi/YeMLWL22}. The segmentation model uses the classical encoder-decoder architecture with a Temporal Context Transformer (TCT) connected at the end of the encoder, the TCT has an encoder-decoder architecture with a timing component combining the characteristics of image sequences. The output features enter the decoder to predict the segmentation masks.
The detection model is a two-stage target detection model, the features are extracted from the encoder and then passed to the temporal context transformer, a feature pyramid network, and a region proposal network. The final predictions are post-processed with the soft-NMS algorithm to remove overlapping bounding boxes.

\newcommand{\jc}{{lswang\underline{ }xmu }}
\subsection*{Team: \jc} 

Team \jc suggested a model that utlized the newest transformer-based network, Polyp-PVT, which separated the polyp lesion in images~\cite{DBLP:conf/isbi/WangMMW22}. The major improvement was the boundary representation enhancement, achieved through the boundary aggregation gate. Specifically, at the end of the pyramid vision transformer, they added the gate to enhance the coarsely embedded features. To increase the diversity and data amount, a series of augmentation functions were adopted, including vertical and horizontal flipping, random rotation, and random Gaussian blurring. An AdamW optimizer with the initial learning rate of 1e-4 was utilized to optimize the parameters. The batch size was set to 8, and the model was trained for a total of 120 epochs using a combination of IoU and cross-entropy loss. The training process is achieved on the Pytorch platform using an NVIDIA GeForce RTX 3080 Ti with the memory of 24 GB. 

\subsection*{Team: SDS-RBS (detection)}
The  SDS-RBS team used YOLOv5 models, mainly YOLOv5l6 and YOLOv5x6 as their \textit{detection} models~\cite{DBLP:conf/isbi/YamlahiGTMATBJM22}. Both models were trained over two folds, where each fold was trained with a different combination of augmentations to improve the model generalization, mainly: Hue-Saturation-Value, copy-paste, mixup and mosaic augmentations. The non-Max-Suppression algorithm was applied to the predictions to remove overlapping boxes. The predictions of both models were ensembled using the weighted-boxes-fusion algorithm. To capture the temporal information in the video sequences, a second stage Norfair tracker was implemented during inference. The model post-processed the final bounding boxes by reducing their sizes by 2\% to exclude some false positives.

\subsection*{Team: SDS-RBS (segmentation)}
For \textit{segmentation} task team SDS-RBS used a heterogeneous model ensemble, combining neural architectures with complementary strength~\cite{DBLP:conf/isbi/TranImYAGTM22}. This was achieved by employing a nnU-Net, which is designed to autonomously adjust its preprocessing and training framework to various datasets, hence providing a robust foundation for segmentation, a Hierarchical Multi-Scale Attention Network (HM-ANet) with Region Mutual Information Loss, which combines predictions of multiple scales for better prediction performance on polyps of varied size, and an ensemble of Efficient-UNets, one of which is equipped with an internal GRU-layer to process temporal information. The HM-ANet weight in the ensemble was increased and was expected to perform better on minimal or huge polyps, as it operates on higher resolutions. A majority vote then yields the final prediction. A post-processing step considers the structural similarity of neighboring images and allows single-model predictions despite non-majority, allowing more temporally coherent results. All models used public polyp datasets from CVC and Etis-Larib as part of their training. The final ensemble achieves a DSC score of 0.74, improving the best-performing single model HM-ANet by 0.04.

\subsection*{Team: UCU ML Lab}
Team UCU ML Lab chose MaskFormer \cite{cheng2021per} as the main model for their segmentation method~\cite{DBLP:conf/isbi/KokshaikynaYD22}. In the model, they transitioned from cross-entropy loss to focal loss to address the class imbalance in classification, reduced the number of queries from 100 to 50, lowered the dimensionality of FC layers from 2048 to 24, and decreased the pixel size from 256 to 64. Moreover, a standard convolution ResNet50 backbone was used instead of SWIN because transformer backbones have poor performance in datasets with few samples, and this was proven in our experiments as well. Normalization coefficients were recalculated for the PolypGen dataset. Furthermore, the model used two types of losses together during training: one for classifying the data and another for the binary mask for each predicted segment. The binary loss is a linear combination of focal, and dice losses \cite{lin2017focal}. They also experimented with boundary loss \cite{2021boundary}. However, the experiments didn't show any positive impact of boundary loss for polyp segmentation. The team used the segmentation mask outputs to report their detection results for this challenge. We have not included this team for detection task ranking. 

\subsection*{Team: WürzVision}
For the training, Team: WürzVision used two additional open-source datasets: Kvasir-SEG and UN Colonoscopy Video Database~\cite{DBLP:conf/isbi/KrenzerSHP22}. The data collection includes 1000 polyp frames, 1071 masks, and bounding boxes. This dataset comprises 49,136 polyp frames from 100 distinct polyps. All models were trained on an NVIDIA QUADRO RTX 8000. The method is based on concepts from Gong et al. \cite{gong2021temporal}, employing deep convolutional neural networks (CNNs) and temporal data to enhance current polyp detection techniques. It illustrates a detection method that employs analogous features from various frames, in conjunction with temporal attention, to forecast the ultimate polyp detections for a new frame. To reduce generalization error, data augmentation is employed, and supplementary open-source data is incorporated for training.

\begin{table*}[t!]
\footnotesize
\centering
\caption{Summary of the detection and segmentation tasks undertaken by the participating teams for the EndoCV2022 challenge. Methods are presented along with their characteristics and the rationale for their selection as considered by the teams. Some codes from teams are accessible for replication purposes.}
\begin{tabular}{lllllllll}
\toprule
\bf Team Name &\bf Algorithm & \bf Backbone &\bf Nature  &\bf \begin{tabular}[l]{@{}c@{}}
    Choice  \\ Basis\end{tabular} & \bf \begin{tabular}[l]{@{}c@{}}
    Data  \\ Aug.\end{tabular} & \bf 
    Loss  & \bf Opt.&\bf Code\\  \midrule  \midrule
    \\
\multicolumn{8}{l}{\textbf{Task I:} Polyp detection}\\
\hline
IMed & \begin{tabular}[l]{@{}l@{}}Two Stage +\\ Detector\\\end{tabular} & \begin{tabular}[l]{@{}l@{}}ResNest +\\ TCT\\\end{tabular} & Aggregation & \begin{tabular}[l]{@{}l@{}@{}} Context \\ Speed \end{tabular} & Yes& \begin{tabular}[l]{@{}l@{}@{}} Cascade \\ Loss  \end{tabular} & \begin{tabular}[l]{@{}l@{}@{}} SGD + \\ Warmup  \end{tabular} &  N/A\\ 
SDS\_RBS & YOLO & \begin{tabular}[l]{@{}l@{}}YOLOv5l6\\ YOLOv65x6\\\end{tabular}  & Ensemble & \begin{tabular}[l]{@{}l@{}}Accuracy\\ Speed\\\end{tabular}  & Yes & CIoU & SGD  & N/A\\ 

WürzVision  & TRA &\begin{tabular}[l]{@{}l@{}}ResNet50\end{tabular}  & \begin{tabular}[l]{@{}l@{}}Temporal\\ Attention\end{tabular}  & F1-Score & Yes & IoU & SGD & \href{https://github.com/Adrian398/AI-JMU-Polyp-Detection-Challenge}{\nolinkurl{[d2]}} \\ 

\multicolumn{8}{l}{\textbf{Task II:} Polyp segmentation}\\
\hline 
Arrah\_htic  & XP-Net  &  U-Net  & \begin{tabular}[l]{@{}l@{}}Knowledge\\ distillation\\\end{tabular} & \begin{tabular}[l]{@{}l@{}}Accuracy\\ Speed\\\end{tabular} & Yes & \begin{tabular}[l]{@{}l@{}}DSC \\ Tversky\end{tabular}   & Adam & \href{https://github.com/Ragu2399/XP-NET}{\nolinkurl{[s2]}} \\

He\_HIK (Quan He) & \begin{tabular}[l]{@{}l@{}}Improved-\\ STCN\\\end{tabular} & \begin{tabular}[l]{@{}l@{}}ResNet18\\ ResNet50\\\end{tabular} & \begin{tabular}[l]{@{}l@{}}Temporal\\ tracking\end{tabular} & Dice Score & Yes & CE & Adam & N/A \\ 

IMed & UNet & \begin{tabular}[l]{@{}l@{}}PraNet +\\ TCT\\\end{tabular} & Aggregation & \begin{tabular}[l]{@{}l@{}@{}} Context \\ Speed \end{tabular}& Yes &\begin{tabular}[l]{@{}l@{}@{}}CE Loss\\ Focal Loss\\ Dice Loss\end{tabular} & \begin{tabular}[l]{@{}l@{}@{}} SGD + \\ Warmup  \end{tabular} & N/A\\ 

\jc & Polyp-PVT &  PVT  & Attention & Dice Score & Yes & \begin{tabular}[l]{@{}l@{}}IoU\\ CE\end{tabular}   & AdamW  & N/A
\\ 

SDS\_RBS  & \begin{tabular}[l]{@{}l@{}@{}}UNet\\ HRANet\\ nnUnet\end{tabular} & \begin{tabular}[l]{@{}l@{}@{}}EfficientNet B5\\ HRNet\\ UNet\end{tabular} & Ensemble & Dice Score & Yes &\begin{tabular}[l]{@{}l@{}@{}}DSC\\ RMI\\ BCE\end{tabular}  & Adam& N/A \\ 

UCU ML Lab & MaskFormer &  \begin{tabular}[l]{@{}l@{}}ResNet50\\DETR\end{tabular}   & Attention &Dice Score  & Yes & \begin{tabular}[l]{@{}l@{}}CE\\ Binary mask\end{tabular}   & SGD  & 
\href{https://github.com/ucuapps/Modified-MaskFormer-for-Polyps-Segmentation}{\nolinkurl{[s4]}} \\

\\ 

\bottomrule
\multicolumn{9}{l}{\footnotesize{CIoU: Complete intersection over union; RMI: Region Mutual Information; ATSS: Adaptive Training Sample Selection}}\\
\multicolumn{9}{l}{\footnotesize{YOLO: You Only Look Once; SGD: Stochastic Gradient Descent; }}\\
\multicolumn{9}{l}{\footnotesize{TCT: Temporal Context Transformer}}\\
\multicolumn{9}{l}{\footnotesize{TRA: Temporal ROI Align; N/A: Not available}}\\
\multicolumn{9}{l}{\footnotesize{CE: Cross entropy; DSC: Dice similarity coefficient; IoU: Intersection over union; PVT: Pyramid Vision Transformer }}
\end{tabular}
\label{table:challenge_summary_detection_segmentation}
\end{table*}
\section*{Results}
%
%
The following is an overview of the results of the leading teams that participated in the EndoCV2022 competition. Each of these teams has taken part in detection tasks, segmentation tasks, or both. The section provides quantitative and qualitative results for each team.
\subsection*{Detection task}

\begin{figure}[t!]
    \centering\includegraphics[width=\textwidth]{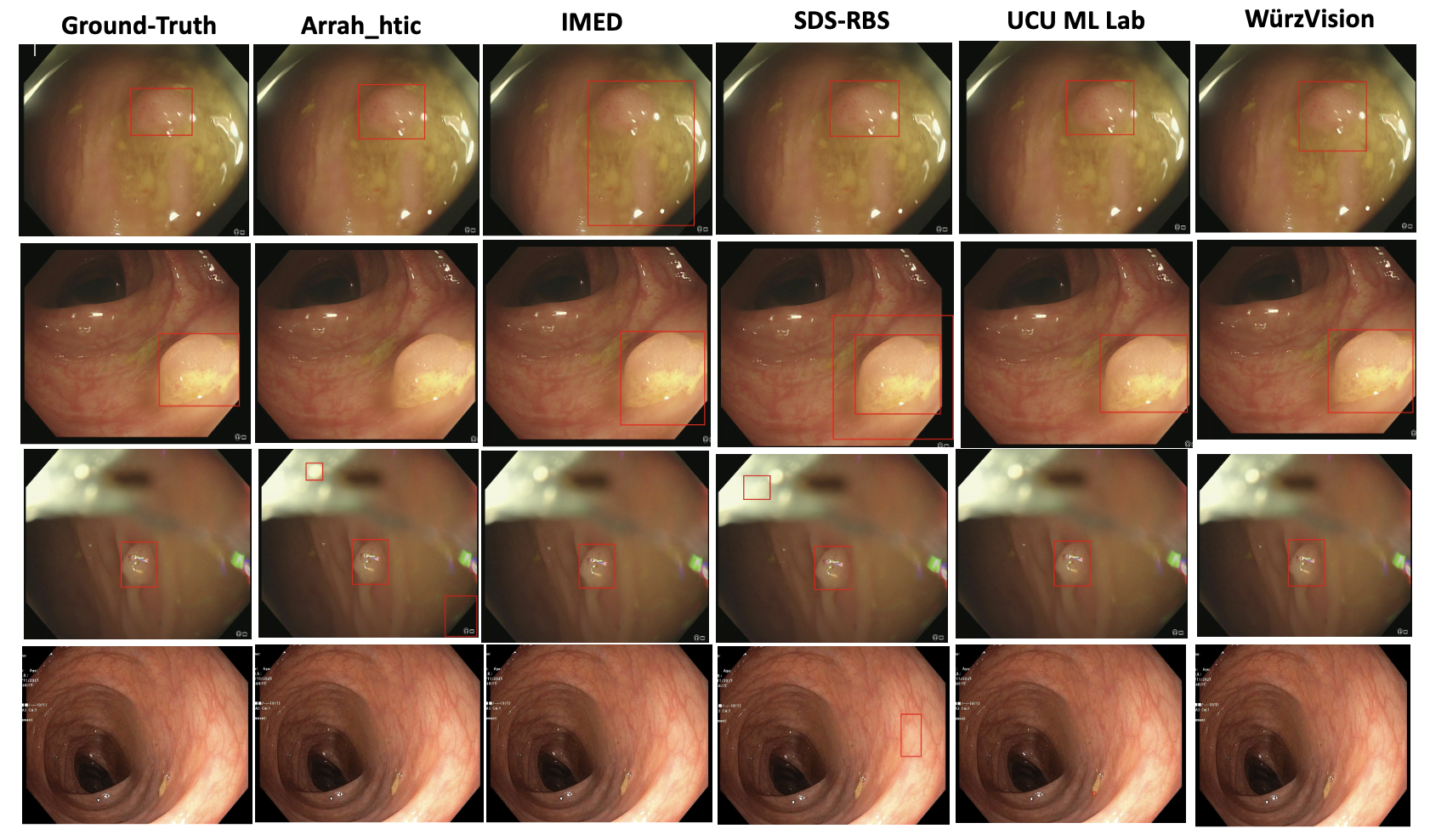}
    \caption{Qualitative results for the detection task displaying both the ground truth and the outcomes from the participating teams. The first three rows illustrate variations in polyp size and location, while the fourth row features a negative sample with no polyp present.}
    \label{fig:det}
\end{figure}

This section presents the outcomes of the top 5 participating teams in the detection task. Table \ref{detectiontable} shows the average precision (AP) values calculated at different intersection over union (IoU) thresholds (i.e. specifically 25\%, 50\%, and 75\%). Team \textit{SDS-RBS} demonstrated superior performance compared to other teams in terms of \textit{AP} values at various IoU thresholds with \textit{$AP_{mean}$} of 0.334 by employing a model that brought together YoloV516 and YoloV5x6 with Norfair tracking for temporal refinement. The approach proposed by the UCU ML Lab team ranks second with a $AP_{mean}$ of 0.1462. However, it did not surpass team \textit{WürzVision}, which scored higher \textit{$AP_{50}$} (0.2705) and \textit{$AP_{large}$} (0.2877). Moreover, Figure \ref{fig:det} illustrates frame samples with ground truth and bounding boxes generated by the methods proposed by the teams. The first two rows show clear polyp samples, emphasizing the precise identification of the methods proposed by the participating teams. The polyps in the third row have blurry surroundings, indicating more challenging scenarios for the methods proposed by the participating teams. The fourth row includes images without polyps, which serve as a test for negative samples against the proposed models. The figures show that several teams exhibited outstanding results by accurately generating bounding boxes that corresponded to the ground truth. However, other teams showed instances of false positives in their images, which can be related to the complicated features inherent in endoscopic imaging.

\subsection*{Segmentation task}
\begin{figure}
    \centering
    \includegraphics[width=\textwidth]{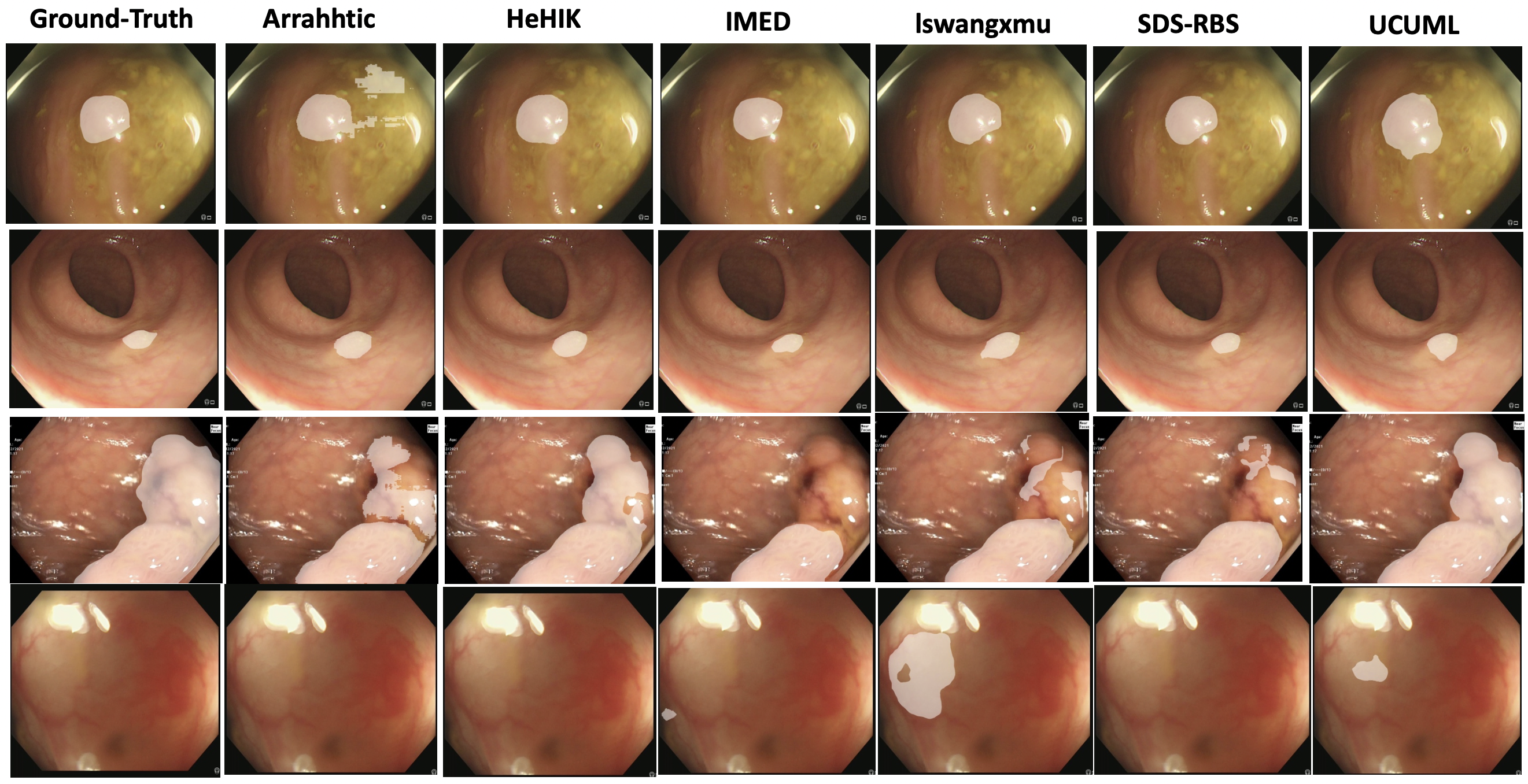}
    \caption{Qualitative results for the segmentation task displaying both the ground truth and the outcomes from the participating teams. The first three rows illustrate variations in polyp size and location, while the fourth row features a negative sample with no polyp present, i.e., presence of artefact.}
    \label{fig:enter-label}
\end{figure}

Table \ref{segtable} displays the segmentation results for the participating teams which includes Jaccard Index (JC), Dice Similarity (DSC), F2-Measure, Postive Predicted Value (PPV), Recall (Rec), Accuracy (Acc), Normalized Surface Distance (NSD), MASD and Hausdorff Distance (d\_hfd). The approaches proposed by the team \textit{HeHIK} and \textit{lswangxmu} demonstrated superior performance compared to the baseline methods employed by other teams in terms of JC(>0.72), DSC(>0.76), F2(>0.76), Recall(>0.80) and Accuracy(>0.99). Nevertheless, \textit{SDS-RBS} achieved the highest PPV with a value of (0.936), followed by \textit{HeHIK} with a value of (0.897), demonstrating their ability to identify more regions with positive predictions. Furthermore, Figure \ref{fig:enter-label} presents qualitative samples of segmentation results. The first three rows display polyps of various sizes, including those with complex features. The fourth row, alternatively, presents samples of polyp-free images, providing a thorough assessment of segmentation performance across various scenarios. As shown, the first two images exhibit similar performance throughout the teams, as they accurately segmented the polyps. The third image displays a complex polyp structure that was inadequately segmented by certain teams. Furthermore, some false positives were detected by the models on polyp-free images.

\subsection*{Runtime and team ranking}
The computational resources, training time, and inference time are summarized in Table \ref{runtime}. In the polyp detection task, team \textit{IMED} acquired the shortest training time of 1 hour, utilizing a 2X Tesla V100 32G GPU, along with a detection rate of 72 frames per second (FPS). Team \textit{SDS-RBS} required a 5-hour training time on an NVIDIA RTX 3090, while team \textit{WürzVision} needed a longer training time of 12 hours with an extended 2 hours of fine-tuning. This illustrates the model complexity and fine-tuning methodologies that improve accuracy but also increase computational time. For the polyp segmentation task, the difference in training time durations among the teams reflects the diversity of the models proposed. The model proposed by team \text{lswang} exhibited a long training duration of 18 hours but achieved a low inference time of 1.25 FPS, in contrast to teams \textit{IMED} and \textit{Arrah\_htic}, which showed better GPU utilization with a shorter training time of 12 hours, with higher inference times of 120 FPs and 104 FPs, respectively. The variance in inference times between the teams highlights the tradeoff between computational efficiency and segmentation accuracy, depending on GPU time and network architecture. 

The overall ranking was determined primarily by the algorithmic performance of the method proposed by each team, followed by the computational speed. For the polyp detection task, team \textit{SDS\_RBS} obtained the best overall rank, although it had the worst inference performance compared to the other two teams; however, it had an outstanding algorithmic detection performance. Team \textit{WürzVision} gained the second rank with a reasonable balance between speed and performance. Finally, team \textit{IMED} ranked third due to its algorithmic efficacy, although they had the best training time. The final results for the segmentation task indicate that both teams \textit{HE\_HIK} and \textit{Iswang} achieved the best performance, maintaining a balance between accuracy and speed efficiency. With a consistent trade-off between segmentation accuracy and inference time, team \textit{HE\_HIK} demonstrated a superior accuracy performance, although with a slower inference time, compared to team \textit{Iswang}. Following in the mid-range ranks, teams \textit{IMED} and \textit{SDS\_RBS} demonstrated a competitive performance, however, they did not excel in segmentation performance or inference time. In contrast, team \textit{Arrah\_htic} had an outstanding inference time, but with a very low segmentation performance. Similarly, the performance of team UCUML in both measures led to the lowest rank. These findings highlight the importance of algorithmic robustness compared to speed performance. 
 \begin{table}[t!]
 \small
 \centering
 \caption{Team results for the {detection task} with average precision AP computed at IoU thresholds 50 (AP$_{50}$), 75 (AP$_{75}$), and $[0.50:0.05:0.95]$ mean AP (AP$_{mean}$). Size wise AP values are also presented. Top-two values for each
metrics are highlighted in bold.\label{sp_detection_table_1}}
 \begin{tabular}{l|l|ll|lll}
 \toprule
& \multicolumn{3}{|c|}{\bf{Average precision, AP}} &  \multicolumn{3}{|c}{{\bf AP across scales}} \\
 \bf{Teams/Method} & \multicolumn{3}{|c|}{} &\multicolumn{3}{|c}{} \\
  & \multicolumn{1}{|c}\bf{AP}$_{mean}$      & \bf{AP}{$_{50}$}   & \bf{AP}{$_{75}$}   & \multicolumn{1}{|c}{}\bf{AP}$_{small}$ & \bf{AP}$_{medium}$ & \bf{AP}$_{large}$ \\ \hline \midrule  
 {Arrahhtic}$^*$ & 0.117 & 0.208 & 0.093 & 0.001 & 0.095 & 0.274\\
{IMED} & 	0.116 &	0.185&	0.141 & 0.000 & 0.102 & 0.256\\
{SDS-RBS} & \textbf{0.334}  & \textbf{0.481} & \textbf{0.367} & \textbf{0.038}& \textbf{0.338}& \textbf{0.677} \\
{UCU ML Lab}$^*$& \textbf{0.146} & 0.266 & \textbf{0.143} & \textbf{0.006} & \textbf{0.184} & 0.184\\
{WürzVisio}n& 0.131&\textbf{0.271}&0.104& 0.000& 0.115& \textbf{0.287}\\
\bottomrule
\multicolumn{6}{l}{*: teams utilising segmentation masks for detection task}
%
 \end{tabular}

\label{detectiontable}
\end{table}
\begin{table*}[t!]
\caption{ Team results for the polyp segmentation methods proposed by the participating teams. Top-two values for each
metric are highlighted in bold.}
\footnotesize
\centering
\begin{tabular}{l|c|c|c|c|c|c|c|c|c}
\hline
\textbf{Teams}        & \textbf{JC}$\uparrow$          & \textbf{DSC}$\uparrow$         & \textbf{F2} $\uparrow$          & \textbf{PPV}$\uparrow$          & \textbf{Rec}$\uparrow$          & \textbf{Acc}$\uparrow$          & \textbf{NSD}$\uparrow$          & \textbf{MASD} $\downarrow$           & \textbf{d\_{hfd}}$\downarrow$          \\ \hline
\hline

{Arrahhtic}    & 0.438$\pm$\scriptsize0.42          & 0.481$\pm$\scriptsize0.44          & 0.495$\pm$\scriptsize0.45          & 0.696$\pm$\scriptsize0.39          & 0.635$\pm$\scriptsize0.44          & 0.989$\pm$\scriptsize0.01          & 0.357$\pm$\scriptsize0.39          & 68.091$\pm$\scriptsize144.90         & 0.388$\pm$\scriptsize0.22          \\ \hline
{HeHIK}        & \textbf{0.720$\pm$\scriptsize0.35} & \textbf{0.765$\pm$\scriptsize0.35} & \textbf{0.766$\pm$\scriptsize0.36} & \textbf{0.897$\pm$\scriptsize0.21} & \textbf{0.807$\pm$\scriptsize0.33} & \textbf{0.995$\pm$\scriptsize0.01} & 0.675$\pm$\scriptsize0.35          & 12.496$\pm$\scriptsize59.07          & 0.301$\pm$\scriptsize0.22          \\ \hline
{iMed}         & 0.522$\pm$\scriptsize0.42          & 0.568$\pm$\scriptsize0.43          & 0.560$\pm$\scriptsize0.43          & 0.860$\pm$\scriptsize0.28          & 0.601$\pm$\scriptsize0.43          & 0.989$\pm$\scriptsize0.02          & 0.443$\pm$\scriptsize0.40          & 32.282$\pm$\scriptsize103.65         & 0.392$\pm$\scriptsize0.28          \\ \hline
{lswangxmu}    & \textbf{0.747$\pm$\scriptsize0.34} & \textbf{0.787$\pm$\scriptsize0.34} & \textbf{0.796$\pm$\scriptsize0.35} & 0.893$\pm$\scriptsize0.21         & \textbf{0.843$\pm$\scriptsize0.32} & \textbf{0.995$\pm$\scriptsize0.01} & \textbf{0.715$\pm$\scriptsize0.35} & \textbf{8.917$\pm$\scriptsize38.75}  & \textbf{0.282$\pm$\scriptsize0.21} \\ \hline
{SDS-RBS} & 0.718$\pm$\scriptsize0.37          & 0.754$\pm$\scriptsize0.37          & 0.744$\pm$\scriptsize0.38          & \textbf{0.936$\pm$\scriptsize0.19} & 0.764$\pm$\scriptsize0.36          & 0.995$\pm$\scriptsize0.01          & \textbf{0.683$\pm$\scriptsize0.37} & \textbf{11.008$\pm$\scriptsize42.92} & 0.287$\pm$\scriptsize0.22          \\ \hline
{UCUML}        & 0.500$\pm$\scriptsize0.41          & 0.550$\pm$\scriptsize0.43          & 0.556$\pm$\scriptsize0.43          & 0.606$\pm$\scriptsize0.43          & 0.794$\pm$\scriptsize0.34          & 0.986$\pm$\scriptsize0.03          & 0.448$\pm$\scriptsize0.38          & 53.947$\pm$\scriptsize140.81         & \textbf{0.253$\pm$\scriptsize0.15} \\ 
\hline
\end{tabular}
\label{segtable}
\end{table*}
\begin{table*}[t!]
\caption{Summary of the GPU configuration utilized by each participating team, training time in hours and inference time in frame per second (fps). Teams with longer post-processing are not included in ranking. Ranking for detection is based on \textbf{AP}$_{mean}$ and that for segmentation is based on DSC. }
\footnotesize
\centering
\begin{tabular}{l|c|c|c|c|c|c}
\hline
\textbf{Teams} & \textbf{GPU} & \textbf{Training} & \textbf{Inference} & \textbf{R}$_{time}$ & \textbf{R}$_{algo.}$ & \textbf{R}$_{overall}$\\
 &  & \textbf{(hrs)} & \textbf{(fps)} & & &  \\
\hline
\hline
\multicolumn{4}{l}{\textbf{Task I: Polyp detection}}\\
\hline
 {IMED} & 2$\times$ Tesla V100 32G & 1 &72& 1 & 4& 3 \\
{SDS\_RBS}& 1 NVIDIA RTX3090&5 & 9.10 &  3 & 1 & \textbf{1}\\
{WürzVision} &NVIDIA QUADRO RTX 8000 & 14 & 24 & 2 & 2 & 2\\
\hline
\multicolumn{4}{l}{\textbf{Task II: Polyp Segmentation}}\\
\hline
{Arrah\_htic} & NVIDIA RTX 3090& 12& 120 & 1 & 6 & 6\\
{He\_HIK}& NVIDIA TESLA V100 GPU &12 & 9 &  3& 2 &  \textbf{1}\\
 {IMED} & 2$\times$ Tesla V100 32G & 1 &104& 2 & 4& 3\\
{lswang}& 1$\times$ NVIDIA 3080& 18  & 1.25 & 5 & 1 & 2\\
{SDS\_RBS} &NVIDIA RTX 2080 Ti & N/A& 0.71& 6 & 3 & 4\\
{UCU ML Lab} & NVIDIA GeForce GTX TITAN X& 6 &8.33& 4 & 5& 5\\ 
\bottomrule
\multicolumn{5}{l}{\footnotesize{\textbf{R}$_{overall}$ = 0.25$\times$\textbf{R}$_{time}$ + 0.75$\times$\textbf{R}$_{algo.}$}}\\
 
\end{tabular}
\label{runtime}
\end{table*}

\section*{Discussion}
This study emphasizes the utilization of temporal information available in endoscopic sequences (colonoscopy videos) for the development of deep learning models for polyp detection and segmentation tasks. The hypothesis is to explore and understand whether focus on frame-by-frame dependencies can enhance the model generalizability and resilience across various imaging settings, in contrast to many previous works that primarily utilize static image-based datasets. Incorporating sequence data from six different centers in this data science challenge provided a comprehensive evaluation of baseline approaches and attempts to overcome the limited generalizability associated with single-center methodologies \cite{ali2024assessing}. The results presented in this paper demonstrate that the use of temporal reasoning modules (i.e. \textit{LSTM, transformers, \& temporal context networks}) significantly improves the performance of detection and segmentation tasks in colonoscopy. For example, based on the results of the detection task, the \textit{SDS-RBS} team proposed an ensemble model, complemented by a temporal tracking phase, that achieved the highest average precision (AP$_{mean}$ = 0.334), underscoring the importance of temporal post-processing in reducing false positives and improving consistency between frames. The improved results were attained from integrating the Norfair tracker, which utilized inter-frame correlation to ensure constant detection throughout video sequences.

Similarly, the segmentation models proposed by the best performing teams, such as \textit{lswangxmu} and \textit{HeHIK}, employed strategies that leverage video sequences. Team \textit{lswangxmu} obtained a DSC of 0.787 by utilizing a transformer based model \textit{Polyp-PVT} with data augmentation, while team \textit{HeHIK} achieved DCS of 0.765 with improved STCN for effective  temporal tracking that propagate information across temporal intervals. These results highlight the importance of integrating temporal relations, which reflect the dynamic nature of video captured during clinical procedures, by surpassing conventional frame-based architectures.

Regardless of these promising outcomes, several limitations remain. Most of the proposed models focused on incorporating temporal information by addressing short-term dependencies between sequenced frames, excluding long-term relations across entire video frames. Including long-term relationships can improve performance by reducing false negatives while improving tracking abilities\cite{wang2025improving}. Moreover, the evaluation focused mainly on the detection and segmentation of polyps, ignoring their characterization and classification, which is important in clinical decision-making \cite{johnson2023colorectal, jain2025enhancing}. In addition, a persistent concern is related to the handling of endoscopic artifacts \cite{ali2021deep}, such as specular regions, blurry areas, etc., which were misclassified as anomalies. Figure \ref{fig:det} shows an example of false positive detection in a smoking region by team \textit{SDS\_RBS} while the model proposed by team \textit{lswangxmu} produced a false detection in a specular region. Moreover, the implementation of state-of-the-art methods in standard clinical environments remains uncertain, as approaches in the literature have not been thoroughly evaluated in challenging out-of-distribution contexts crucial for complex polyp detection and segmentation tasks \cite{toman2025espnet}.
\section*{Conclusion}
This paper provides a comprehensive analysis of deep learning techniques for polyp detection and segmentation from sequence data developed by leading contributors in the EndoCV2022 challenge. Our main goal was to assess the effect of training and testing with sequence colonoscopy frames on model performance. We analyzed the collaborative efforts of various teams to address polyp detection and segmentation using sequence data from six different centers. Teams using sequential frames with some incorporation of temporal information showed improved accuracy and robustness in polyp detection and segmentation compared to those that only used individual frames. In summary, this study highlights the importance of temporal modeling in endoscopic video analysis with a focus on enhancing algorithmic generalization and transparency.  Bridging the gap between research and clinical will require more multi-center collaborations, robust validation processes, and explainable AI models to develop reliable and real-time decision support systems.

\bibliography{sample}
\end{document}

%% file: titleAuthors.tex
%
%
%

\title{A multi-center analysis of deep learning methods for video polyp detection and segmentation}
\author[1]{Noha Ghatwary}
\author[2]{Pedro Chavarrias Solano}
\author[1]{Mohamed Ramzy Ibrahim}
\author[3]{Adrian Krenzer}
\author[3]{Frank Puppe}
\author[4]{Stefano {Realdon}}
\author[4]{Renato {Cannizzaro}}
\author[5]{Jiacheng Wang}   
\author[5]{Liansheng Wang} 
\author[6]{Thuy Nuong Tran} 
\author[6]{Lena Maier-Hein} 
\author[6]{Amine Yamlahi} 
\author[6]{Patrick Godau}
\author[7]{Quan He}
\author[7]{Qiming Wan}
\author[8]{Mariia Kokshaikyna} 
\author[8]{Mariia Dobko} 
\author[9]{Haili Ye} 
\author[9]{Heng Li} 
\author[10]{Ragu B} 
\author[10]{Antony Raj} 
\author[11]{Hanaa Nagdy} 
\author[12]{Osama E Salem}
\author[13]{James E. East} 
\author[14]{Dominique {Lamarque}}
\author[15,16]{Thomas de Lange} 
\author[2,*]{Sharib Ali}
\affil[1]{Computer Engineering Department, Arab Academy for Science and Technology,Egypt}
\affil[2]{AI in Medicine and Surgery Group, School of Computer Science, University of Leeds, Leeds, LS2 9JT, UK}
\affil[3]{Department of Artificial Intelligence and Knowledge Systems, University of Würzburg, Germany}
\affil[4]{CRO Centro Riferimento Oncologico IRCCS, Aviano, Italy}
\affil[5]{Department of Computer Science at School of Informatics, Xiamen University}
\affil[6]{Div. Intelligent Medical Systems, German Cancer Research Center (DKFZ), Heidelberg, Germany}
\affil[7]{Hangzhou Hikvision Digital Technology Co.,ltd, Hangzhou, China}
\affil[8]{The Machine Learning Lab, Ukrainian Catholic University, Lviv, Ukraine}
\affil[9]{Research Institute of Trustworthy Autonomous Systems, Southern University of Science and Technology, China}
\affil[10]{Healthcare Technology Innovation Centre, Chennai, India}
\affil[11]{Internal Medicine Department, College of Medicine, Arab Academy for Science and Technology, Egypt}
\affil[12]{Faculty of Medicine, University of Alexandria, Alexandria, Egypt}
\affil[13]{Translational Gastroenterology Unit, John Radcliffe Hospital, University of Oxford, Oxford, UK}
\affil[14]{Universit{\'e} de Versailles St-Quentin en Yvelines, H{\^o}pital Ambroise Par{\'e}, France}
\affil[15]{Department of Molecular and Clinical Medicine, Sahlgrenska Academy, University of Gothenburg, Sweden}
\affil[16]{Department of Medicine, Geriatrics and Emergencies, Sahlgrenska University Hospital-Mölndal, Gothenburg Västra Götaland Region, Sweden}
\affil[*]{corresponding author: Sharib Ali (s.s.ali@leeds.ac.uk)}

%% file: abstract.tex
\begin{abstract}
Colonic polyps are well-recognized precursors to colorectal cancer (CRC), typically detected during colonoscopy. However, the variability in appearance, location, and size of these polyps complicates their detection and removal, leading to challenges in effective surveillance, intervention, and subsequently CRC prevention. The processes of colonoscopy surveillance and polyp removal are highly reliant on the expertise of gastroenterologists and occur within the complexities of the colonic structure. As a result, there is a high rate of missed detections and incomplete removal of colonic polyps, which can adversely impact patient outcomes. Recently, automated methods that use machine learning have been developed to enhance polyps detection and segmentation, thus helping clinical processes and reducing missed rates. These advancements highlight the potential for improving diagnostic accuracy in real-time applications, which ultimately facilitates more effective patient management. Furthermore, integrating sequence data and temporal information could significantly enhance the precision of these methods by capturing the dynamic nature of polyp growth and the changes that occur over time. To rigorously investigate these challenges,  data scientists and experts in gastroenterology collaborated to compile a comprehensive dataset that spans multiple centers and diverse populations collected from six distinct colonoscopy systems. This initiative aims to underscore the critical importance of incorporating sequence data and temporal information in the development of robust automated detection and segmentation methods. Using a crowd-sourced endoscopic computer vision challenge, we strive to advance the field of polyp detection and improve clinical outcomes for patients undergoing colonoscopy.  This study evaluates the applicability of deep learning techniques developed in real-time clinical colonoscopy tasks using sequence data, highlighting the critical role of temporal relationships between frames in improving diagnostic precision. We analyze the outcomes of two best-performing teams for the detection task and six best-performing teams for the segmentation task. Our research findings suggest that the teams incorporated the temporal relationships between frames of the proposed methods to achieve acceptable detection and segmentation results.

\end{abstract}